\documentclass[sigconf]{acmart}

\usepackage{tabularx}
\usepackage{needspace}
\usepackage{placeins}

\acmConference[WWW '27]{The ACM Web Conference 2027}{May 10--14, 2027}{Dublin, Ireland}
\acmYear{2027}
\copyrightyear{2027}
\setcopyright{none}
\acmDOI{}
\acmISBN{}
\title{ColaBench: Evaluating Memory-Guided Knowledge Acquisition in Personal Agents}
\renewcommand{\shorttitle}{ColaBench: Memory-Guided Knowledge Acquisition}

\author{Wenjie Zhang}
\email{wenjie.zhang@stu.jiangnan.edu.cn}
\affiliation[obeypunctuation=true]{
  \institution{Jiangnan University}, \city{Wuxi}, \country{China}
}

\author{Zhichao Shi}
\email{shizhichao22s@ict.ac.cn}
\affiliation[obeypunctuation=true]{
  \institution{School of Advanced Interdisciplinary Sciences, UCAS}, \city{Beijing}, \country{China}
}
\affiliation[obeypunctuation=true]{
  \institution{State Key Lab of AI Safety, Institute of Computing Technology, CAS}, \city{Beijing}, \country{China}
}
\affiliation[obeypunctuation=true]{
  \institution{DataArc Tech Ltd.}, \country{China}
}
\affiliation[obeypunctuation=true]{
  \institution{IDEA Research, International Digital Economy Academy}, \city{Shenzhen}, \country{China}
}

\author{Xuhui Jiang}
\correspondingauthor
\authornote{Primary contact: Xuhui Jiang.}
\email{jiangxuhui@idea.edu.cn}
\affiliation[obeypunctuation=true]{
  \institution{DataArc Tech Ltd.}, \country{China}
}
\affiliation[obeypunctuation=true]{
  \institution{IDEA Research, International Digital Economy Academy}, \city{Shenzhen}, \country{China}
}

\author{Xiaojun Wu}
\email{xwu647@connect.hkust-gz.edu.cn}
\author{Cehao Yang}
\email{cyang289@connect.hkust-gz.edu.cn}
\affiliation[obeypunctuation=true]{
  \institution{DataArc Tech Ltd.}, \country{China}
}
\affiliation[obeypunctuation=true]{
  \institution{IDEA Research, International Digital Economy Academy}, \city{Shenzhen}, \country{China}
}
\affiliation[obeypunctuation=true]{
  \institution{The Hong Kong University of Science and Technology (Guangzhou)}, \city{Guangzhou}, \country{China}
}

\author{Chengjin Xu}
\email{xuchengjin@idea.edu.cn}
\affiliation[obeypunctuation=true]{
  \institution{DataArc Tech Ltd.}, \country{China}
}
\affiliation[obeypunctuation=true]{
  \institution{IDEA Research, International Digital Economy Academy}, \city{Shenzhen}, \country{China}
}

\author{Xiaoning Song}
\correspondingauthor
\email{x.song@jiangnan.edu.cn}
\affiliation[obeypunctuation=true]{
  \institution{Jiangnan University}, \city{Wuxi}, \country{China}
}
\renewcommand{\shortauthors}{Zhang et al.}

\begin{document}
\begin{abstract}

Personal agents need to draw on long-term interaction history to retrieve and apply external knowledge suited to a user's current request.
Even an answer grounded in relevant documents can be unsuitable when it overlooks the user's circumstances.
This creates two connected challenges: using historical conditions to identify the knowledge needed for a new request, and applying that knowledge to the user's situation.
Existing benchmarks advance memory recall, personalized retrieval, and memory-guided action, while the joint evaluation of these two challenges remains underexplored.
We introduce ColaBench, a synthetic benchmark in health and daily living, with 349 tasks across 10 Profiles and a knowledge base of 13,033 documents.
Its Evidence Graph connects personal conditions to their disclosures across public interaction sessions, records prior document exposure, and links answer requirements to personal and newly acquired document evidence.
We also introduce MemKB, which uses state summaries to inform queries and rule application, and memory--document links to retrieve knowledge associated with the user's history.
MemKB achieves 86.25\% Accuracy with an average of 4.07 interaction rounds per task.
Task-level analysis finds larger gains on tasks integrating information across multiple historical Facts; error analysis further identifies unsupported personal assumptions and rule misapplication as obstacles to using retrieved evidence.
ColaBench provides traceable tasks and an evaluation framework for studying how memory guides knowledge acquisition and rule application.
\end{abstract}
\maketitle

\section{Introduction}
\begingroup

Personal agents must draw on long-term interaction history to handle users' evolving requests~\mbox{\cite{zhong2024memorybank,packer2023memgpt}}.
Users often leave prior experiences, preferences, and constraints unstated~\mbox{\cite{jiang2025personamem}}, while the current task may require new external knowledge~\mbox{\cite{lewis2020rag}}.
Consider a user asking whether they qualify for a caregiver respite grant after mentioning, in separate conversations, that they care for a relative with Alzheimer's disease and live alone.
The agent must find the grant's eligibility rules and connect the co-residence requirement to the earlier disclosure of living alone (Table~\ref{tab:case}).
Overlooking that connection can turn relevant information into unsuitable advice.
Reliable personalization therefore depends on how history guides knowledge acquisition and how the acquired knowledge is applied to the user.

These tasks pose two connected challenges.
\textbf{Knowledge acquisition} requires recovering the historical conditions that determine what to search for, even when the current request leaves them implicit.
\textbf{Rule application} requires matching retrieved prerequisites, exceptions, and procedures to known personal circumstances while preserving what remains unknown.
The process can move in both directions: history shapes the search, and a newly retrieved rule can make an earlier disclosure relevant.
When memory and knowledge are retrieved separately, the agent must establish these connections through its queries and reasoning.

Existing benchmarks address important parts of this process.
LoCoMo and LongMemEval provide a foundation for evaluating memory through cross-session question answering, temporal reasoning, and state updates~\cite{maharana2024locomo,wu2025longmemeval}.
TREC iKAT brings user context, query reformulation, and external retrieval into the answering process~\cite{aliannejadi2024ikat}.
Mem2ActBench and MemoryArena examine how memory guides subsequent tool use and interdependent actions~\cite{shen2026mem2actbench,he2026memoryarena}.
A remaining gap concerns the joint evaluation of how disclosed personal conditions guide the acquisition and application of knowledge not yet encountered in the interaction.
This requires tasks that trace what the user has disclosed, what documents the agent has already encountered, and which personal and knowledge premises support each required conclusion.
These dependencies make it possible to examine where the connection between history and knowledge succeeds or breaks down.

We introduce \textbf{ColaBench}, a benchmark built around these dependencies.
Its \textbf{Evidence Graph} connects personal Facts to the public messages that disclose them, records document exposure during interaction, and links subsequent answer requirements to their personal and knowledge sources.
Its document links guide knowledge exploration, and its disclosure and exposure records constrain staged, multi-agent task construction to disclosed personal conditions and supporting documents not previously exposed in the history.
The resulting benchmark comprises 349 enhanced and validated Items across 10 Profiles and a knowledge base of 13,033 documents.
The graph makes each task's evidence dependencies explicit, supporting evaluation of knowledge acquisition and rule application and diagnosis of whether errors arise from missing personal conditions, missing knowledge, or their use in the answer.

We introduce MemKB, a memory-guided knowledge retrieval method that connects historical context to knowledge candidates.
A user state summary brings dispersed personal conditions into query formulation and subsequent judgment.
MemKB encodes memory text, establishes memory--document links, and fuses documents associated with retrieved memories with direct semantic and keyword retrieval results for the current query.
The answering agent combines retrieved memories and document evidence to assess which rules apply to the user's circumstances.

MemKB achieves 86.25\% Accuracy, compared with 79.94\% for Mem0+Hybrid using the same underlying memories, a gain of 6.30 percentage points, while reducing average interaction rounds from 4.44 to 4.07.
The gain is larger on tasks requiring multiple historical Facts: 9.35 percentage points, compared with 4.96 on single-Fact tasks.
Case and error analyses show how personal assumptions and rule misapplication can affect answers even when relevant evidence is available.
These findings motivate evaluating memory through the knowledge it helps agents acquire and the user-specific judgments it supports.

Our contributions are as follows:
\begin{itemize}
  \item \textbf{Evidence Graph--driven benchmark synthesis.} We develop a staged, multi-agent synthesis framework in which accumulated Fact disclosure and knowledge exposure constrain subsequent Item construction, producing the traceable ColaBench benchmark and its evaluation protocol.
  \item \textbf{Memory-guided knowledge retrieval.} We introduce MemKB, which combines state summaries and memory--document links to guide knowledge retrieval and support user-specific rule application.
  \item \textbf{Empirical analysis of memory utility.} Through task-structure groupings, component ablations, and evidence traces, we characterize memory utility in knowledge acquisition and rule application: gains on tasks integrating multiple historical Facts and failures involving omitted personal conditions, unsupported assumptions, and rule misapplication after knowledge access.
\end{itemize}
\par\endgroup
\section{Related Work}
\paragraph{Long-Term Memory and Task Execution}
Memory architectures organize interaction history for subsequent decisions. Generative Agents combine experience retrieval and reflection to support planning~\cite{park2023generative}; MemoryBank updates user memories over time~\cite{zhong2024memorybank}, and MemGPT manages information across memory tiers~\cite{packer2023memgpt}. Mem0 supports memory extraction, consolidation, and retrieval~\cite{chhikara2025mem0}. A-MEM builds evolving links between memory notes~\cite{xu2025amem}, while Graphiti, the graph engine underlying Zep, maintains temporal entity and fact relations~\cite{rasmussen2025zep}.

LoCoMo generates long-term conversations from profiles and temporal event graphs, with human review, to evaluate question answering, event summarization, and multimodal dialogue generation~\cite{maharana2024locomo}. LongMemEval covers cross-session reasoning, temporal information, and state updates, and analyzes memory systems at the indexing, retrieval, and reading stages~\cite{wu2025longmemeval}. Mem2ActBench examines how memory supports subsequent tool selection and argument specification~\cite{shen2026mem2actbench}, and MemoryArena evaluates experience reuse across interdependent multi-session tasks~\cite{he2026memoryarena}. ColaBench studies subsequent actions through knowledge search, evidence reading, and rule application, tracing how recovered user state supports each required conclusion.

\paragraph{Personalized Conversational Search}
LaMP studies retrieval of user-specific records for personalized classification and generation~\cite{salemi2024lamp}. PersonaMem evaluates profile tracking and response selection as users' circumstances evolve~\cite{jiang2025personamem}. TREC iKAT is a closely related benchmark for personalized knowledge acquisition. Personal textual knowledge bases specify user backgrounds, and conversational tasks require systems to use this personal context. Evaluation covers passage relevance as well as the relevance, completeness, grounding, and naturalness of generated answers~\cite{aliannejadi2024ikat}. ColaBench records personal-state disclosure and knowledge exposure across simulated interactions. Its tasks trace evidence from public user statements through rule premises to required judgments, enabling evaluation of history-guided knowledge retrieval and rule application.

\paragraph{Evidence-Level Diagnosis}
Retrieval-augmented generation uses external documents to generate answers~\cite{lewis2020rag}. HippoRAG combines knowledge graphs and Personalized PageRank to integrate evidence across passages for multi-hop question answering~\cite{gutierrez2024hipporag}. ALCE evaluates answer correctness and the quality of supporting citations~\cite{gao2023alce}. RAGChecker uses fine-grained claims to diagnose retrieval coverage and evidence use during generation, helping identify cases where relevant documents have been retrieved but required content is omitted~\cite{ru2024ragchecker}. We use necessary answer requirements as diagnostic units and connect them to the sources of personal conditions and the applicability conditions of rules. This enables us to examine whether the required evidence was obtained and correctly applied.
\section{Evidence-Graph-Based Construction}
\label{sec:construction}
\begin{figure*}[t]
\centering
\includegraphics[width=\textwidth]{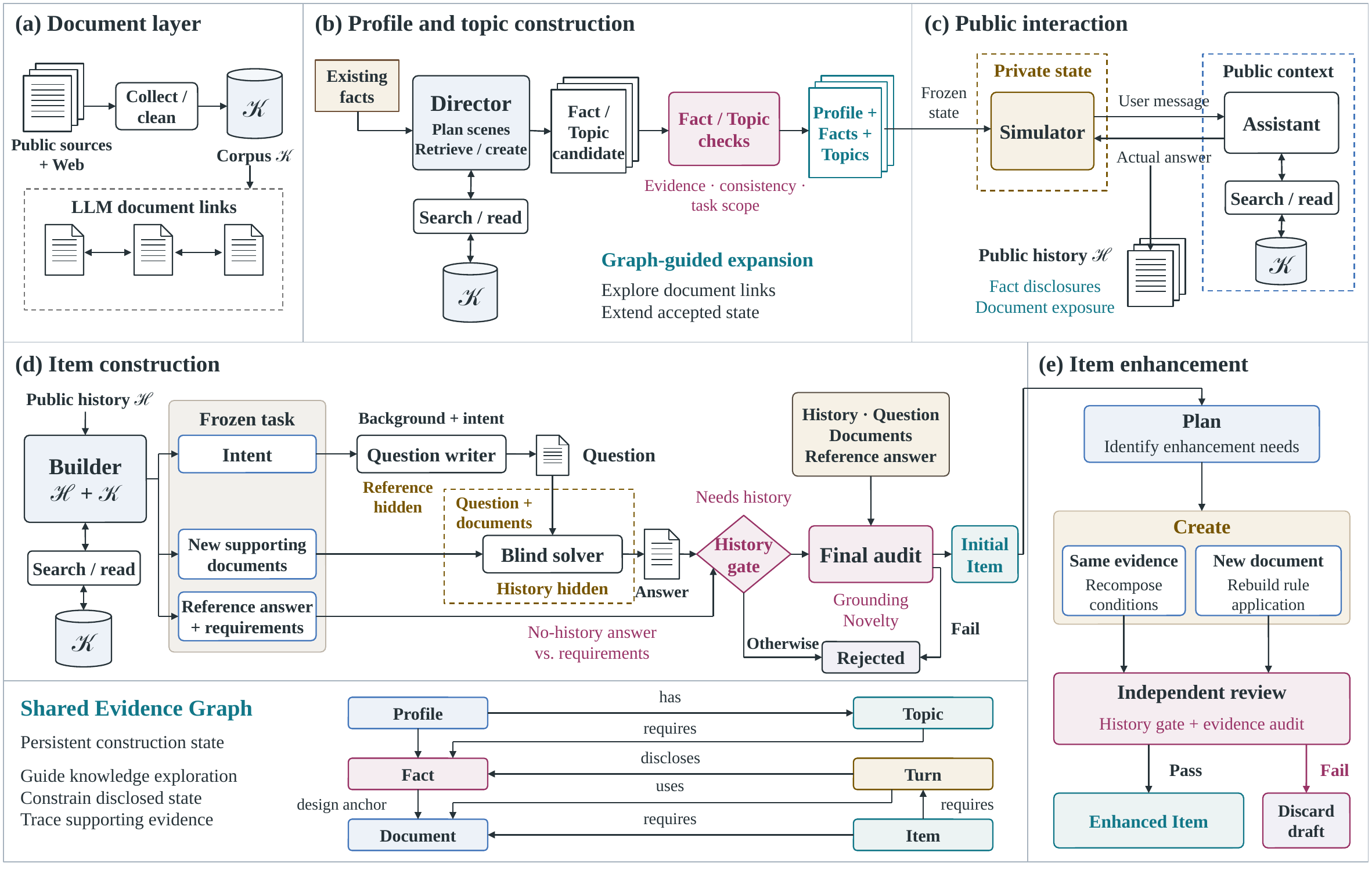}
\caption{Plan--Create--Review construction: (a) document collection; (b) grounded Profiles and Topics; (c) public interaction; (d) history-dependent Item construction; and (e) Item enhancement and independent validation. The growing Evidence Graph guides knowledge exploration and constrains generation and review through recorded Fact disclosures, document exposure, and supporting evidence.}
\Description{Panels a to c show document collection, grounded Profile and Topic construction, and public interaction with separate private and public inputs. Panel d shows initial Item construction: a question writer cannot see the reference answer, a blind solver cannot see history, and a history gate plus final audit select Items. Beside it, panel e analyzes each initial Item, creates an enhanced draft using either the original evidence or an alternative supporting document, and independently validates history dependence and evidence support. Passing drafts become enhanced Items; failed drafts are discarded. The bottom Evidence Graph guides knowledge exploration, constrains disclosed state and document novelty, and supports evidence validation across stages.}
\label{fig:construction}
\end{figure*}

ColaBench uses a \textbf{Plan--Create--Review} multi-agent framework with stage-specific roles and information access to ground synthetic personal backgrounds, interaction histories, and tasks in retrieved documents. An \textbf{Evidence Graph} $\mathcal G=(V,E)$ begins with documents and their relationships and expands throughout construction. The graph incorporates Document, Profile, Fact, Topic, Turn, and Item nodes. Document links guide knowledge exploration, while Fact--Topic relations organize user states and interaction tasks. Records of fact disclosure and document exposure determine which personal conditions and new knowledge are available for later Items. Figure~\ref{fig:construction} shows the roles and information boundaries across document preparation, the three generation stages, and Item enhancement.

\subsection{Task Definition}
\label{sec:task}
The evaluation tasks examine how an agent uses personal conditions from history to acquire and apply external knowledge. We denote the task setting by $x=(q,\mathcal H,\mathcal K;\mathcal M)$, where $q$ is the current request, $\mathcal H$ is the preceding public interaction history, $\mathcal K$ is the frozen knowledge base, and $\mathcal M$ denotes the evaluation metrics defined in Section~\ref{sec:evaluation}.

\begin{table}[tbp]
\centering
\caption{A ColaBench task (paraphrased).}
\label{tab:case}
\small
\begin{tabularx}{\columnwidth}{@{}p{0.18\columnwidth}X@{}}
\toprule
Component & Description\\
\midrule
Request & Am I eligible for an HFC caregiver respite care grant?\\
\addlinespace[2pt]
History & In a meal-preparation conversation, the user states that they \textbf{live alone}. In another exchange, they mention caring for a relative with Alzheimer's disease.\\
\addlinespace[2pt]
Rule & The care recipient must \textbf{live at home with the applicant}.\\
\addlinespace[2pt]
Judgment & Living alone \textbf{does not satisfy} the co-residence requirement; other undisclosed conditions remain unknown.\\
\bottomrule
\end{tabularx}
\end{table}

Personal conditions can guide knowledge search, and newly retrieved rules can prompt the agent to revisit history.  Table~\ref{tab:case} illustrates this interaction through a grant's co-residence requirement.

\subsection{Stage 0: Building the Document Layer}
We collect public health information, drug labels, and open textbooks from screened public sources through APIs, XML files, and web pages. After text cleaning, the base corpus contains 11,069 documents. Each document forms an initial Document node in the Evidence Graph, retaining its text and source information. Appendix~\ref{app:corpus-sources} details the corpus sources and processing.

Starting from seed documents, an LLM retrieves and reads related documents to establish edges grounded in specific knowledge relationships, storing a description and weight for each relation. We further search, fetch, and filter web pages on relevant topics, incorporating the resulting documents into the corpus and graph. This document layer supplies knowledge and exploration links for Profile generation and Item construction. Web pages discovered during these stages are also added as Document nodes.

\subsection{Stage I: Grounding Profiles and Facts}
To construct internally consistent users with varied life circumstances, a Profile construction agent, the Director, plans new scenarios from the existing profile and consults retrieved documents to create Facts describing personal conditions and Topics specifying interaction tasks. An independent review agent checks consistency, relevance, and task completeness against the existing state and source material. Accepted content is added to the Evidence Graph. Documents ground the synthetic scenarios in relevant contexts and knowledge, while links among Profiles, Facts, Topics, and Documents support continued profile development across life domains and provide evidence for subsequent interactions.

\subsection{Stage II: Growing Public Interaction Traces}
Interactions between a user Simulator and an Assistant gradually disclose personal conditions as tasks unfold. The Simulator has access to the frozen Profile and Topic, whereas the Assistant responds using the public conversation and retrieved knowledge. The Simulator follows up or switches Topics in response to the Assistant's actual replies, producing a continuous Trace driven by task progress. Accepted interactions enter the Evidence Graph as Turn nodes, recording Fact disclosures and document exposure to establish the history available for subsequent construction.

\subsection{Stage III: Constructing Items}
To construct an initial Item, the Builder starts from disclosed personal conditions and seeks new knowledge whose application requires those conditions, establishing the task and its answer evidence. An independent question-writing role then formulates a natural request from the public background and task intent, leaving the necessary personal context in history. A history-removal test checks whether the task can be completed without the public interaction history and whether an unmet answer requirement depends on missing historical conditions. Subsequent review assesses source support, knowledge novelty, and continuity with the preceding interaction. Accepted Items enter the Evidence Graph with links to the required Facts, historical Turns, and supporting Documents, providing a traceable account of how personal conditions affect the application of new knowledge. Construction details and evidence annotations appear in Appendix~\ref{app:protocol}.

\subsection{Item Enhancement and Validation}
An enhancement agent examines each initial Item's historical conditions and knowledge rules to develop decision tasks that combine multiple personal conditions (Figure~\ref{fig:construction}(e)). With the public interaction history fixed, it chooses one of two enhancement directions: reorganizing condition dependencies within the original evidence or replacing the supporting document with a retrieved candidate from the same Topic. It updates the question, reference answer, and answer requirements accordingly. An independent History Gate and evidence review assess history dependence, source support, and the combination of conditions. Appendix~\ref{app:enhancement} specifies the acceptance and validation procedures.
\subsection{Dataset Overview}
\label{sec:dataset-overview}
ColaBench contains 10 Profiles, 349 Items, and 13,033 documents, as summarized in Table~\ref{tab:dataset-summary}. A multi-Fact task cites at least two annotated Facts; a task citing multiple sessions draws its annotated historical evidence from at least two sessions. Appendix~\ref{app:dataset-structure} provides detailed structural statistics; Table~\ref{tab:state-rule-transfer} in Appendix~\ref{app:shared-background} illustrates tasks sharing personal backgrounds.

\begin{table}[htbp]
\centering
\caption{Structural statistics of ColaBench.}
\label{tab:dataset-summary}
\small
\begin{tabularx}{\columnwidth}{@{}Xr@{}}
\toprule
Statistic & Value \\
\midrule
Profiles / Sessions / Messages & 10 / 21 / 1,198 \\
Facts / Topics / Items & 52 / 600 / 349 \\
Base / Added web documents & 11,069 / 1,964 \\
Items with multiple Facts & 107 \\
Items citing multiple sessions & 153 \\
\bottomrule
\end{tabularx}
\end{table}
\section{MemKB: Memory-Linked Retrieval}
\label{sec:method}
\begin{figure*}[t]
\centering
\includegraphics[width=\textwidth]{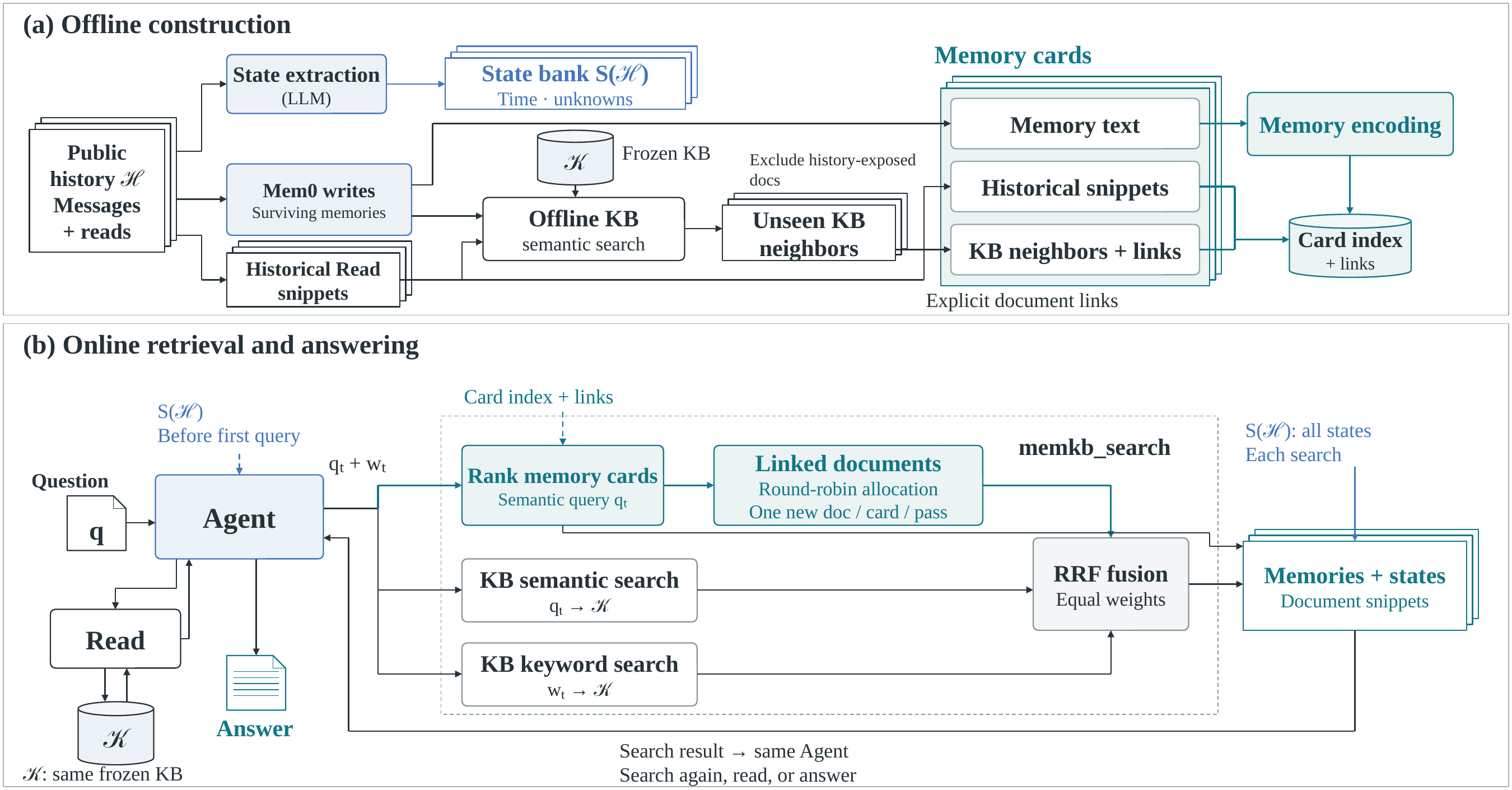}
\caption{ MemKB construction and answering. (a) Public history yields state summaries and memory cards linked to historical Read documents and unexposed KB neighbors. (b) States inform the first query; round-robin link expansion and equal-weight RRF combine memory-linked, semantic, and keyword candidates. Search returns memories, states, and document snippets.}
\Description{Public interaction history produces a compact state bank and Mem0-derived memory cards. Card vectors encode memory text, and explicit document links connect cards to historical Read documents and offline KB neighbors. The state bank is supplied before the first query. A query ranks cards, whose linked documents are allocated round-robin, alongside semantic and keyword KB matches. Matched memory text with source identifiers, the complete state bank, and RRF-fused document snippets return to the agent, which can search again, read documents, or answer.}
\label{fig:method}
\end{figure*}

{
MemKB connects two decisions: which knowledge the user's request requires and how the retrieved rules apply to their circumstances. MemKB supplies a user state summary before query generation and retrieves knowledge through explicit links between memory cards and documents. It then fuses these candidates with documents retrieved directly for the current query (Figure~\ref{fig:method}).
}

{
\subsection{Memory Cards and Document Links}
Given a public interaction history $\mathcal H$ and a frozen knowledge base $\mathcal K$, we use Mem0~\cite{chhikara2025mem0} to build personal memory incrementally and retain its write records. These records link each historical interaction to its surviving memory IDs, forming cards $m_i$ with source message IDs and normalized embeddings of their memory text. The document identifiers and start positions in the interaction's public Read records allow the builder to recover short snippets from the frozen KB and establish document links.

For each card, the builder queries the full KB using memory text together with historical document titles and snippets. It retains top semantic candidates not previously exposed in the Profile's history as offline document links. These neighbors and historical Read documents form the card's linked document set, connecting memory to both previously used and unexplored knowledge. Neighbor discovery uses public interaction history and the full frozen KB, while reference-document annotations are withheld from the builder. Encoder settings and budgets appear in Appendix~\ref{app:run-configuration}.

A language model extracts a compact state bank $S(\mathcal H)$ from the complete public interaction history, preserving who each condition concerns, its temporal context, unresolved conditions, and source message IDs. The complete $S(\mathcal H)$ is added to the system prompt before the first model call, allowing personal conditions to inform the initial query. Each search returns the same complete state bank to support interpretation of retrieved evidence and assessment of rule applicability. Query-relevant cards add memory text, source-message IDs, and knowledge links.

\subsection{Memory-Linked Knowledge Retrieval}
\label{sec:memory-linked-retrieval}
Hybrid retrieves documents directly from the KB through semantic and keyword channels. At retrieval step $t$, the agent uses the current request $q$ and information acquired so far to generate a semantic query $q_t$ and a keyword set $w_t$. Vector similarity and term matching produce candidate lists $L_{\mathrm{sem}}$ and $L_{\mathrm{key}}$, respectively, whose rankings are then fused. In baselines with separate memory and knowledge retrieval, a dedicated tool returns personal memory, which influences knowledge retrieval through the queries formulated by the agent.

MemKB adds a third retrieval channel that transfers query relevance from memory cards to linked documents. The same $q_t$ retrieves cards by normalized inner product. Following card rank order, a round-robin procedure takes one document from each card per pass, skipping documents already in the candidate list, until the candidate budget is reached or all cards are exhausted. This produces $L_{\mathrm{mem}}$. Distributing candidate slots across relevant cards allows knowledge associated with different historical conditions to enter the candidate set.

Reciprocal rank fusion (RRF)~\cite{cormack2009rrf} combines the memory-linked candidates with Hybrid's two direct retrieval channels, incorporating both associations with personal history and direct matches to the current query:
\begin{equation}
 s(d)=\sum_{c\in\{\mathrm{mem},\mathrm{sem},\mathrm{key}\}}\frac{\mathbf{1}[d\in L_c]}{\kappa+\operatorname{rank}_{L_c}(d)}.
\end{equation}
Here, $L_c$ is the candidate list for channel $c$, $\kappa$ is a rank-smoothing constant, and ranks start at 1. Each document contributes at most once per channel, and all three channels receive equal weight. The fused ranking determines the returned documents, each accompanied by a relevant snippet, its document ID, and a line range.

A call to \texttt{memkb\_search} returns memory text and source identifiers from matched cards, $S(\mathcal H)$, and the fused document snippets.

\subsection{From User State to Rule Application}
The same answering agent compares the conditions specified by retrieved rules with the user's state and memory records, continuing to search or call Read as needed. Its answering prompt includes a source-attribution constraint: distinguish the speaker, actor, and event status in memory records, and ground personal communications and decisions in the records that support them. When the source or event status is unspecified, the agent is instructed to preserve that uncertainty. This constraint guides the agent to retain the provenance of personal conditions when applying rules.
}
\section{Experimental Setup}
\label{sec:setup}
\subsection{Methods and Experimental Settings}
We pair Mem0, Graphiti, and A-MEM with a general-purpose retrieval-augmented generation component, Hybrid, and compare them with No Memory+Hybrid and MemKB. Hybrid combines semantic and keyword retrieval over the KB through RRF~\cite{cormack2009rrf}.

The Mem0+Hybrid baseline uses Mem0's non-graph memory~\cite{chhikara2025mem0}. Graphiti+Hybrid retrieves facts about entities and their relations extracted from conversations~\cite{rasmussen2025zep}, while A-MEM+Hybrid uses memory evolution and neighbor expansion~\cite{xu2025amem}. These three methods build memory from the same public user and assistant messages. MemKB additionally uses public records of historical document access to build a memory-linked retrieval channel and supplies a state summary before the first query.

The main comparison uses all 349 tasks from 10 Profiles and the same frozen KB across methods. All methods can make tool calls for up to four rounds per task, followed by a final answering round. Knowledge search returns five documents by default, with a maximum of 20. Read returns at most 100 lines and 3,000 tokens per call.

{
All methods use DeepSeek v4.1 Flash for answering. GLM-5.3 Flash evaluates each answer in a separate call with low reasoning effort. Model parameters appear in Appendix~\ref{app:run-configuration}, and the shared evaluation protocol appears in Appendix~\ref{app:strict-accuracy-prompts}.
\par}

\subsection{Evaluation Metrics}
\label{sec:evaluation}
\textbf{Accuracy} measures whether an answer correctly completes the current task. A passing answer must address all explicitly requested subquestions and provide correct core conclusions and key actions, including the relevant personal conditions, applicability requirements, and limitations. The evaluator checks factual errors and omissions by asking whether correcting them would change the conclusion, action, timing, applicability, or safety. Errors and omissions that affect task completion lead to failure; incidental differences are recorded separately. Grounded equivalent wording and implicit use of personal conditions are accepted, while unknown facts require the necessary conditional branches.  Evaluation inputs combine the full public history, reference answer, requirements, historical and task-supporting documents, and each method's actual Search and Read returns. Metric and input details appear in Appendix~\ref{app:metric-details}.

\textbf{Read Recall} measures coverage of annotated new-support documents $G_i$ by documents $D_i^{\mathrm{read}}$ for which Read returns nonempty text before the final answer. Each task has one reference document. Across $N$ evaluated tasks,
\begin{equation}
\mathrm{Read\ Recall}=\frac{1}{N}\sum_{i=1}^{N}\frac{|G_i\cap D_i^{\mathrm{read}}|}{|G_i|}.
\end{equation}
Search snippets and information already present in memory can also support an answer directly, so we analyze document coverage alongside Accuracy.

\textbf{Avg. Interaction Rounds} is the mean number of agent rounds, including the final answering round.

We additionally assess five answer-quality dimensions. Each dimension receives a separate blinded GLM-5.3 Flash (low) call using the same reference inputs and candidate-visible evidence as Accuracy. Scores range from 1 to 5, with higher scores indicating better quality; Appendix~\ref{app:answer-dimensions} defines the dimensions.
\section{Results and Analysis}
\label{sec:results}

\subsection{Answer Quality and Knowledge Access}
On 349 tasks (Table~\ref{tab:main-results}), MemKB achieves 86.25\% Accuracy, exceeding Mem0+Hybrid by 6.30 percentage points despite close Read Recall (45.56\% versus 44.70\%). This gap motivates examining how answers use personal conditions and retrieved evidence. No Memory+Hybrid achieves 6.59\% Accuracy versus Graphiti+Hybrid's 80.80\%, illustrating the importance of recovering personal context. MemKB exceeds A-MEM+Hybrid by 4.87 percentage points and uses 0.35 and 0.37 fewer interaction rounds than A-MEM+Hybrid and Mem0+Hybrid, respectively.

\begin{table}[t]
\centering
\caption{Main comparison on 349 tasks. Acc. and Read Recall are percentages; rounds include the final answer.}
\label{tab:main-results}
\label{tab:cost-results}
\small
\setlength{\tabcolsep}{4pt}
\begin{tabular}{@{}lrrr@{}}
\toprule
Method & Acc. $\uparrow$ & Read Recall $\uparrow$ & Avg. Rounds $\downarrow$\\
\midrule
No Memory+Hybrid~\cite{cormack2009rrf} & 6.59 & 37.25 & 4.85\\
Graphiti+Hybrid~\cite{rasmussen2025zep} & 80.80 & 41.83 & 4.58\\
A-MEM+Hybrid~\cite{xu2025amem} & 81.38 & 39.26 & 4.42\\
Mem0+Hybrid~\cite{chhikara2025mem0} & 79.94 & 44.70 & 4.44\\
MemKB (ours) & 86.25 & 45.56 & 4.07\\
\bottomrule
\end{tabular}
\end{table}

{

\par}

\paragraph{Condition Coverage and Rule Application}
Table~\ref{tab:answer-dimensions} evaluates the five dimensions on 111 tasks from three Profiles. MemKB scores higher across all dimensions, with the largest gains in Condition Coverage and Rule Application (both +0.207). On the 16 tasks answered correctly by MemKB and incorrectly by Mem0+Hybrid, the mean Rule Application gain is 1.563 points.  These results link memory use to identifying needed knowledge and applying it to the user's circumstances.

\begin{table}[t]
\centering
\caption{Model-assessed answer-quality scores on 111 tasks from three Profiles. Each dimension is scored from 1 to 5.}
\label{tab:answer-dimensions}
\small
\setlength{\tabcolsep}{4pt}
{
\begin{tabular}{@{}lrrr@{}}
\toprule
Dimension & Mem0+Hybrid & MemKB & Gain\\
\midrule
State Fidelity & 4.523 & 4.595 & +0.072\\
Condition Coverage & 4.532 & 4.739 & +0.207\\
Evidence Grounding & 4.252 & 4.405 & +0.153\\
Rule Application & 4.396 & 4.604 & +0.207\\
Answer Completeness & 4.586 & 4.766 & +0.180\\
\bottomrule
\end{tabular}}
\end{table}

\begingroup

\begin{figure*}[t]
\centering
\includegraphics[width=\textwidth]{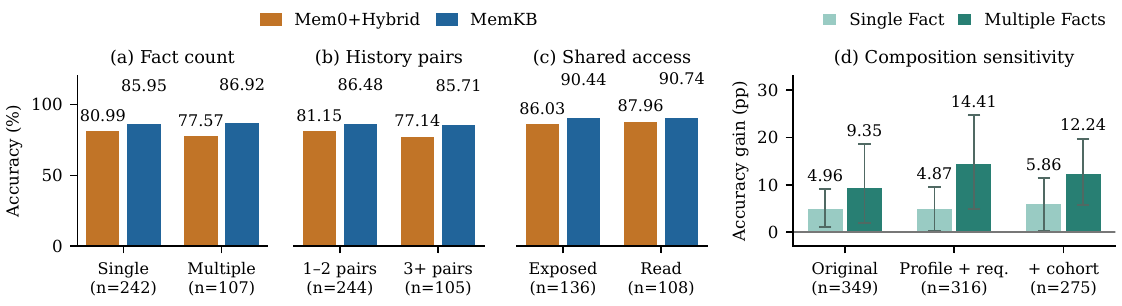}
\caption{Task structure and evidence access on 349 tasks. (a,b) Accuracy by Fact and history-pair counts. (c) Both exposed: both methods receive the reference document through search or Read; both read: through Read. (d) Paired gains under common composition weights with 95\% Profile-cluster bootstrap intervals. pp: percentage points; req.: answer requirements.}
\Description{Four panels show gains by task structure, shared reference access, and task-composition adjustments with confidence intervals.}
\label{fig:experiment-analysis}
\end{figure*}

\subsection{Combining Personal Information}
\paragraph{Personal Conditions and Their History Sources}
We group tasks by the number of annotated Facts and cited user--assistant interaction pairs to examine tasks that combine more personal information. In Figure~\ref{fig:experiment-analysis}(a,b), the Accuracy gain is 4.96 percentage points on single-Fact tasks and 9.35 points on multi-Fact tasks. The gain is 5.33 points on tasks citing one or two interaction pairs, and 8.57 points on tasks citing at least three pairs. These groupings overlap: 91 multi-Fact tasks also cite at least three interaction pairs. MemKB gains more when tasks combine personal information.

\paragraph{Sensitivity to Task Composition}
Stratifying by Profile and answer-requirement count and applying common weights to strata containing both Fact groups yields gains of 4.87 and 14.41 percentage points on 316 tasks (Figure~\ref{fig:experiment-analysis}(d)). Adding construction cohort yields gains of 5.86 and 12.24 points for single-Fact and multi-Fact tasks on 275 tasks. Appendix~\ref{app:composition-sensitivity} details the strata and bootstrap intervals. Larger observed gains across these analyses highlight the value of organizing multiple historical conditions for knowledge selection and rule application.

\begin{table*}[t]
\centering
\caption{Paired cases of personal context and rule application. Responses are paraphrased.}
\label{tab:paired-cases}
\small
\setlength{\tabcolsep}{5pt}
{
\begin{tabularx}{\textwidth}{@{}>{\raggedright\arraybackslash}p{.33\textwidth}>{\raggedright\arraybackslash}X>{\raggedright\arraybackslash}X@{}}
\toprule
Task and decisive premise & Mem0+Hybrid & MemKB\\
\midrule
\textbf{Breastfeeding suitability.} The user has a history of childhood cancer and works in maternal and child health. Applying the guidance requires distinguishing past from current treatment.
& Retrieved memories emphasize teaching experience. The answer claims no record of cancer treatment, omitting personal history relevant to rule applicability. \textcolor[RGB]{180,45,45}{\textbf{Verdict: incorrect.}}
& The initial state summary preserves the cancer history. The answer distinguishes past from current treatment and keeps unknown treatment details conditional. \textcolor[RGB]{32,112,80}{\textbf{Verdict: correct.}}\\
\addlinespace[4pt]
\textbf{Keeping naloxone at home.} The user takes a prescription pain medicine of unspecified class. A personal recommendation depends on opioid use and overdose risk.
& Reads the relevant guidance, then asserts in the answer that the user takes an opioid. The risk discussion proceeds from this unconfirmed premise. \textcolor[RGB]{180,45,45}{\textbf{Verdict: incorrect.}}
& The state summary records a prescription pain medicine. The answer still asserts opioid use, leaving dose and concurrent medicines as the unresolved details. \textcolor[RGB]{180,45,45}{\textbf{Verdict: incorrect.}}\\
\bottomrule
\end{tabularx}
}
\end{table*}

\subsection{Shared Access and Answer Quality}
We examine answer quality on subsets defined by both methods' observed reference-document access (Figure~\ref{fig:experiment-analysis}(c)). The 136 tasks satisfying \emph{Both exposed} are those where each method receives the reference document through search or a nonempty Read response. Accuracy is 86.03\% for Mem0+Hybrid and 90.44\% for MemKB. On the 108 tasks satisfying \emph{Both read}, where each method receives document text through Read, Accuracy is 87.96\% and 90.74\%, respectively. Even with access to the same document, the visible passages, available personal conditions, and their application can differ. Together with the following cases, these answer gaps highlight the role of personal conditions in applying retrieved knowledge.

\subsection{Case Studies: Rule Application}
\label{sec:case-study}
Table~\ref{tab:paired-cases} contrasts recovery of personal history with a shared failure to preserve an unknown condition; Appendix~\ref{app:paired-diagnosis} provides source evidence.

\paragraph{Personal History Beyond the Query Topic}
Despite reading the same contraindication document, Mem0+Hybrid's two memory searches emphasize teaching content, while its answer claims no record of cancer treatment. MemKB's state summary preserves the user's cancer history; its answer distinguishes past from current treatment and keeps unknown treatment and medication details conditional.

\paragraph{Handling Unconfirmed Personal Conditions}
Both methods correctly answer the naloxone-access question but assert opioid use to assess personal risk. MemKB's state summary says ``prescription pain medicine,'' yet its answer assigns an opioid class. Prior discussion of a drug class does not establish personal use: applying a rule requires grounding known conditions and retaining conditional branches for unknown premises.

\endgroup
\subsection[State Summaries and Document Links]{State Summaries and Document Links}
\label{sec:component-interventions}
We evaluate component ablations on 111 tasks from three Profiles. The ablations remove either the state summary or document links, while retaining the answering model, evaluation criteria, and interaction budget in Section~\ref{sec:setup}. The ablation setup is detailed in Appendix~\ref{app:component-ablations}.

Removing state summaries and document links reduces Accuracy by 6.31 and 10.81 percentage points, respectively (Table~\ref{tab:native-link-ablation}).

\begin{table}[htbp]
\centering
\caption{Ablations on 111 tasks from three Profiles; changes from MemKB appear in parentheses (pp or rounds).}
\label{tab:native-link-ablation}
\small
\setlength{\tabcolsep}{3pt}
\begin{tabular}{@{}lrrr@{}}
\toprule
Method & Acc. $\uparrow$ & Read Recall $\uparrow$ & Avg. Rounds $\downarrow$\\
\midrule
MemKB & 91.89 & 42.34 & 4.27\\
w/o States & 85.59\,{\scriptsize($-6.31$)} & 40.54\,{\scriptsize($-1.80$)} & 4.49\,{\scriptsize($+0.22$)}\\
w/o Links & 81.08\,{\scriptsize($-10.81$)} & 36.94\,{\scriptsize($-5.41$)} & 4.41\,{\scriptsize($+0.14$)}\\
\bottomrule
\end{tabular}
\end{table}

\paragraph{Organizing Personal Conditions}
Removing the state summary eliminates state information both before the first query and from search responses, while preserving query-relevant memory text. The first-search hit rate decreases from 36.94\% to 22.52\%, and average interaction rounds increase from 4.27 to 4.49. Organizing these conditions explicitly helps the agent locate relevant knowledge earlier and carry personal conditions into its subsequent judgments.

\paragraph{Connecting Memories to Knowledge}
Removing document links preserves the state summary and returned memory text, with knowledge candidates supplied by semantic and keyword retrieval. Read Recall decreases from 42.34\% to 36.94\%, while the search hit rate over the full interaction increases from 46.85\% to 61.26\%. These divergent trends distinguish search exposure from subsequent reading: with document links, MemKB reads the reference document on more tasks and completes more tasks correctly.

{
 Together, the 111-task ablations support using organized personal conditions and memory-linked knowledge to guide acquisition and ground rule application.
\par}

\subsection{Error Analysis}
\label{sec:error-analysis}
\begin{figure}[!t]
\centering
\includegraphics[width=\columnwidth]{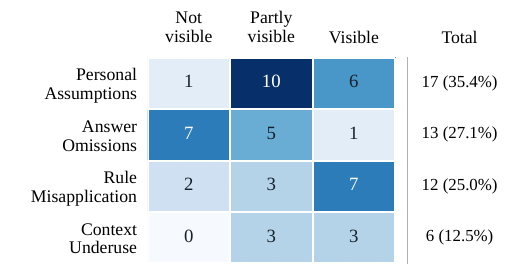}
\caption{Primary error categories and model-assessed evidence visibility for 48 MemKB failures. Cells show counts; row totals include percentages. Other Issues has zero cases.}
\Description{A four-by-three count heatmap. Personal Assumptions has 1 not-visible, 10 partly-visible and 6 visible cases; Answer Omissions has 7, 5 and 1; Rule Misapplication has 2, 3 and 7; Context Underuse has 0, 3 and 3.}
\label{fig:error-analysis}
\end{figure}

We analyze the 48 failed MemKB answers using GLM-5.3 Flash (low), with the complete evaluation references and actual answering traces. Figure~\ref{fig:error-analysis} relates each answer's primary error category to the visibility of its key evidence. Appendix~\ref{app:error-diagnosis} defines the categories, primary-category selection, and evidence visibility.

Personal Assumptions is the largest category. Another Profile's naloxone task illustrates an error during state extraction. The user confirms taking a prescription pain medicine, but the derived state assigns it to the opioid class. The answer then treats opioid-related guidance as personally applicable. This error begins in the memory representation; the case in Section~\ref{sec:case-study} illustrates an unsupported inference during answering. Both show why the source and uncertainty of personal conditions matter when applying rules.

Answer Omissions and Rule Misapplication expose different failure points. A scalp-cooling answer omits a contraindication absent from its retrieved context, whereas a neuropathy answer recommends foot baths despite having read the relevant heat-risk guidance. The former reflects missing decision-relevant evidence; the latter reflects incorrect application of a visible rule to the user.

{
These errors show that memory-guided knowledge tasks require both decision-relevant evidence and a faithful account of the user's conditions when applying it.
\par}

\section{Discussion and Limitations}
\label{sec:discussion}
Method development and comparison use ten Profiles, with component ablations on three selected Profiles. Generalization to new users and the reliability of automated evaluation require independent data and human calibration. The frozen health and everyday-life corpus makes document versions, sources, and reading scope reproducible; generalization to other domains and live Web environments remains to be evaluated. Knowledge comes from real documents, while interactions encode synthetic personal states. Task answers have not been clinically validated.
\section{Conclusion}
ColaBench evaluates memory-guided knowledge acquisition and application through an Evidence Graph linking answer requirements to historical and document evidence. MemKB improves task accuracy with fewer interaction rounds, with ablations supporting the contributions of state summaries and document links. Together, they connect memory evaluation to the knowledge agents acquire and the user-specific judgments they make.

\bibliographystyle{ACM-Reference-Format}
\bibliography{references}

@inproceedings{wu2025longmemeval,
  author = {Di Wu and Hongwei Wang and Wenhao Yu and Yuwei Zhang and Kai-Wei Chang and Dong Yu},
  title = {{LongMemEval}: Benchmarking Chat Assistants on Long-Term Interactive Memory},
  booktitle = {International Conference on Learning Representations},
  year = {2025},
  url = {https://openreview.net/forum?id=pZiyCaVuti}
}

@inproceedings{shen2026mem2actbench,
  author = {Yiting Shen and Kun Li and Wei Zhou and Songlin Hu},
  title = {{Mem2ActBench}: A Benchmark for Evaluating Long-Term Memory Utilization in Task-Oriented Autonomous Agents},
  booktitle = {Proceedings of the 64th Annual Meeting of the Association for Computational Linguistics (Volume 1: Long Papers)},
  year = {2026}, pages = {8173--8190},
  address = {San Diego, California, United States},
  publisher = {Association for Computational Linguistics},
  doi = {10.18653/v1/2026.acl-long.370},
  url = {https://aclanthology.org/2026.acl-long.370/}
}

@inproceedings{maharana2024locomo,
  author = {Adyasha Maharana and Dong-Ho Lee and Sergey Tulyakov and Mohit Bansal and Francesco Barbieri and Yuwei Fang},
  title = {Evaluating Very Long-Term Conversational Memory of {LLM} Agents},
  booktitle = {Proceedings of the 62nd Annual Meeting of the Association for Computational Linguistics (Volume 1: Long Papers)},
  year = {2024}, pages = {13851--13870},
  address = {Bangkok, Thailand},
  publisher = {Association for Computational Linguistics},
  doi = {10.18653/v1/2024.acl-long.747},
  url = {https://aclanthology.org/2024.acl-long.747/}
}

@inproceedings{aliannejadi2024ikat,
  author = {Mohammad Aliannejadi and Zahra Abbasiantaeb and Shubham Chatterjee and Jeffery Dalton and Leif Azzopardi},
  title = {{TREC iKAT 2023}: A Test Collection for Evaluating Conversational and Interactive Knowledge Assistants},
  booktitle = {Proceedings of the 47th International ACM SIGIR Conference on Research and Development in Information Retrieval},
  year = {2024}, publisher = {Association for Computing Machinery},
  address = {New York, NY, USA}, numpages = {11},
  doi = {10.1145/3626772.3657860},
  url = {https://arxiv.org/abs/2405.02637}
}

@misc{ru2024ragchecker,
  author = {Dongyu Ru and Lin Qiu and Xiangkun Hu and Tianhang Zhang and Peng Shi and Shuaichen Chang and Cheng Jiayang and Cunxiang Wang and Shichao Sun and Huanyu Li and Zizhao Zhang and Binjie Wang and Jiarong Jiang and Tong He and Zhiguo Wang and Pengfei Liu and Yue Zhang and Zheng Zhang},
  title = {{RAGChecker}: A Fine-grained Framework for Diagnosing Retrieval-Augmented Generation},
  year = {2024}, eprint = {2408.08067}, archivePrefix = {arXiv},
  url = {https://arxiv.org/abs/2408.08067}
}

@misc{chhikara2025mem0,
  author = {Prateek Chhikara and Dev Khant and Saket Aryan and Taranjeet Singh and Deshraj Yadav},
  title = {{Mem0}: Building Production-Ready {AI} Agents with Scalable Long-Term Memory},
  year = {2025}, eprint = {2504.19413}, archivePrefix = {arXiv},
  url = {https://arxiv.org/abs/2504.19413}
}

@misc{zhang2025qwen3,
  author = {Yanzhao Zhang and Mingxin Li and Dingkun Long and Xin Zhang and Huan Lin and Baosong Yang and Pengjun Xie and An Yang and Dayiheng Liu and Junyang Lin and Fei Huang and Jingren Zhou},
  title = {{Qwen3 Embedding}: Advancing Text Embedding and Reranking Through Foundation Models},
  year = {2025}, eprint = {2506.05176}, archivePrefix = {arXiv},
  url = {https://arxiv.org/abs/2506.05176}
}

@inproceedings{lewis2020rag,
  author = {Patrick Lewis and Ethan Perez and Aleksandra Piktus and Fabio Petroni and Vladimir Karpukhin and Naman Goyal and Heinrich K{\"u}ttler and Mike Lewis and Wen-tau Yih and Tim Rockt{\"a}schel and Sebastian Riedel and Douwe Kiela},
  title = {Retrieval-Augmented Generation for Knowledge-Intensive {NLP} Tasks},
  booktitle = {Advances in Neural Information Processing Systems},
  year = {2020}, volume = {33},
  url = {https://arxiv.org/abs/2005.11401}
}

@inproceedings{cormack2009rrf,
  author = {Gordon V. Cormack and Charles L. A. Clarke and Stefan B{\"u}ttcher},
  title = {Reciprocal Rank Fusion Outperforms Condorcet and Individual Rank Learning Methods},
  booktitle = {Proceedings of the 32nd International ACM SIGIR Conference on Research and Development in Information Retrieval},
  year = {2009},
  url = {https://plg.uwaterloo.ca/~gvcormac/cormacksigir09-rrf.pdf}
}

@inproceedings{xu2025amem,
  author = {Wujiang Xu and Zujie Liang and Kai Mei and Hang Gao and Juntao Tan and Yongfeng Zhang},
  title = {{A-Mem}: Agentic Memory for {LLM} Agents},
  booktitle = {Advances in Neural Information Processing Systems},
  year = {2025}, volume = {38},
  doi = {10.52202/085713-0593},
  url = {https://proceedings.neurips.cc/paper_files/paper/2025/hash/19909c36f51abc4856b4560aff3d36d6-Abstract-Conference.html}
}

@misc{rasmussen2025zep,
  author = {Preston Rasmussen and Pavlo Paliychuk and Travis Beauvais and Jack Ryan and Daniel Chalef},
  title = {{Zep}: A Temporal Knowledge Graph Architecture for Agent Memory},
  year = {2025}, eprint = {2501.13956}, archivePrefix = {arXiv},
  url = {https://arxiv.org/abs/2501.13956}
}

@inproceedings{park2023generative,
  author = {Joon Sung Park and Joseph O'Brien and Carrie Jun Cai and Meredith Ringel Morris and Percy Liang and Michael S. Bernstein},
  title = {Generative Agents: Interactive Simulacra of Human Behavior},
  booktitle = {Proceedings of the 36th Annual ACM Symposium on User Interface Software and Technology},
  year = {2023}, publisher = {Association for Computing Machinery},
  address = {New York, NY, USA}, articleno = {2}, numpages = {22},
  doi = {10.1145/3586183.3606763},
  url = {https://doi.org/10.1145/3586183.3606763}
}

@inproceedings{zhong2024memorybank,
  author = {Wanjun Zhong and Lianghong Guo and Qiqi Gao and He Ye and Yanlin Wang},
  title = {{MemoryBank}: Enhancing Large Language Models with Long-Term Memory},
  booktitle = {Proceedings of the AAAI Conference on Artificial Intelligence},
  year = {2024}, volume = {38}, number = {17}, pages = {19724--19731},
  doi = {10.1609/aaai.v38i17.29946},
  url = {https://ojs.aaai.org/index.php/AAAI/article/view/29946}
}

@misc{packer2023memgpt,
  author = {Charles Packer and Sarah Wooders and Kevin Lin and Vivian Fang and Shishir G. Patil and Ion Stoica and Joseph E. Gonzalez},
  title = {{MemGPT}: Towards {LLMs} as Operating Systems},
  year = {2023}, eprint = {2310.08560}, archivePrefix = {arXiv},
  url = {https://arxiv.org/abs/2310.08560}
}

@inproceedings{salemi2024lamp,
  author = {Alireza Salemi and Sheshera Mysore and Michael Bendersky and Hamed Zamani},
  title = {{LaMP}: When Large Language Models Meet Personalization},
  booktitle = {Proceedings of the 62nd Annual Meeting of the Association for Computational Linguistics (Volume 1: Long Papers)},
  year = {2024}, pages = {7370--7392},
  address = {Bangkok, Thailand},
  publisher = {Association for Computational Linguistics},
  doi = {10.18653/v1/2024.acl-long.399},
  url = {https://aclanthology.org/2024.acl-long.399/}
}

@inproceedings{jiang2025personamem,
  author = {Bowen Jiang and Zhuoqun Hao and Young-Min Cho and Bryan Li and Yuan Yuan and Sihao Chen and Lyle Ungar and Camillo J. Taylor and Dan Roth},
  title = {Know Me, Respond to Me: Benchmarking {LLMs} for Dynamic User Profiling and Personalized Responses at Scale},
  booktitle = {Second Conference on Language Modeling},
  year = {2025},
  url = {https://openreview.net/forum?id=6ox8XZGOqP}
}

@inproceedings{he2026memoryarena,
  author = {Zexue He and Yu Wang and Churan Zhi and Yuanzhe Hu and Tzu-Ping Chen and Lang Yin and Ze Chen and Tong Arthur Wu and Siru Ouyang and Zihan Wang and Jiaxin Pei and Julian McAuley and Yejin Choi and Alex Pentland},
  title = {Benchmarking Agent Memory in Interdependent Multi-Session Agentic Tasks},
  booktitle = {Proceedings of the 43rd International Conference on Machine Learning},
  year = {2026}, volume = {306}, pages = {41975--42005},
  series = {Proceedings of Machine Learning Research}, publisher = {PMLR},
  url = {https://proceedings.mlr.press/v306/he26am.html}
}

@inproceedings{gutierrez2024hipporag,
  author = {Jim{\'e}nez Guti{\'e}rrez, Bernal and Yiheng Shu and Yu Gu and Michihiro Yasunaga and Yu Su},
  title = {{HippoRAG}: Neurobiologically Inspired Long-Term Memory for Large Language Models},
  booktitle = {Advances in Neural Information Processing Systems},
  year = {2024}, volume = {37},
  doi = {10.52202/079017-1902},
  url = {https://proceedings.neurips.cc/paper_files/paper/2024/hash/6ddc001d07ca4f319af96a3024f6dbd1-Abstract-Conference.html}
}

@inproceedings{gao2023alce,
  author = {Tianyu Gao and Howard Yen and Jiatong Yu and Danqi Chen},
  title = {Enabling Large Language Models to Generate Text with Citations},
  booktitle = {Proceedings of the 2023 Conference on Empirical Methods in Natural Language Processing},
  year = {2023}, pages = {6465--6488},
  address = {Singapore}, publisher = {Association for Computational Linguistics},
  doi = {10.18653/v1/2023.emnlp-main.398},
  url = {https://aclanthology.org/2023.emnlp-main.398/}
}
\FloatBarrier
\appendix
\section{Dataset and Construction Protocol}
\label{app:protocol}

\subsection{Corpus Sources}
\label{app:corpus-sources}
Table~\ref{tab:corpus-sources} summarizes the sources of the frozen knowledge base. We extract text from public APIs, XML files, web pages, and open textbooks, removing formatting markup and navigation content while preserving titles, body text, and source URLs. Documents are organized by entry, chapter, or drug-label field. Web additions are deduplicated by body text.

\begin{table}[t]
\centering
\caption{Sources of the frozen knowledge base. Web additions include selected supporting material and related candidate documents.}
\label{tab:corpus-sources}
\small
\begin{tabularx}{\columnwidth}{@{}Xr@{}}
\toprule
Source & Documents\\
\midrule
\href{https://medlineplus.gov/xml.html}{MedlinePlus} & 4,849\\
\href{https://open.fda.gov/apis/drug/label/}{openFDA drug/label} & 2,803\\
\href{https://www.cancer.gov/about-cancer}{National Cancer Institute (NCI)} & 2,371\\
\href{https://www.nichd.nih.gov/health/topics}{NICHD Health Topics} & 818\\
\href{https://odphp.health.gov/myhealthfinder}{MyHealthfinder} & 103\\
\href{https://www.wistechopen.org/open-rn-details}{Open RN nursing textbooks} & 90\\
\href{https://open.lib.umn.edu/physicalactivity/chapter/chapter-1/}{A Guide to Physical Activity} & 24\\
\href{https://www.cdc.gov/diabetes/about/index.html}{CDC health information} & 11\\
\midrule
Base corpus & 11,069\\
Web additions & 1,964\\
Total corpus & 13,033\\
\bottomrule
\end{tabularx}
\end{table}

\subsection{Task Annotations}
\label{app:dataset-structure}
Table~\ref{tab:full-distributions} describes the annotations of the 349 tasks. Single-Fact and multi-Fact tasks cite one Fact and at least two Facts, respectively. A Fact may contain several related personal conditions. Historical citations are counted as user--assistant interaction pairs, and cross-session statistics follow the definition in Section~\ref{sec:dataset-overview}. The 1,015 answer requirements are the criteria used to evaluate Accuracy; their counts are also used in the task-composition analysis in Section~\ref{sec:results}.

\begin{table}[t]
\centering
\caption{Distribution of annotation counts per task. Each cell gives the number of tasks, and each row sums to 349.}
\label{tab:full-distributions}
\small
\setlength{\tabcolsep}{4pt}
\begin{tabular}{lrrrrrr}
\toprule
Count per task & 1 & 2 & 3 & 4 & 5 & 6\\
\midrule
Facts & 242 & 90 & 15 & 2 & 0 & 0\\
Cited turn pairs & 22 & 222 & 79 & 22 & 2 & 2\\
Requirements & 3 & 110 & 171 & 47 & 17 & 1\\
\bottomrule
\end{tabular}
\end{table}

\subsection{History and Evidence Construction}
Public user statements establish personal conditions, and the current request follows all public sessions. The  interaction histories retain tool-call arguments and document navigation information; the full text of documents used in history is available in the frozen knowledge base. Offline memory construction uses the public interaction history and frozen knowledge base. During online execution, the agent receives the current request and obtains relevant information through the states, memories, and tool outputs supplied by each method. Private Facts, reference answers, answer requirements, and reference-document annotations are withheld from the methods.

Completed Topics are sampled without replacement using a fixed random seed, with one initial Item construction attempt per Topic.  Tasks constructed from the frozen corpus use public interaction history and knowledge-base search and read tools.  For web-supported tasks, the Builder uses the current Topic and the source interactions for its relevant Facts to search web pages and obtain supporting text. New supporting documents exclude documents already exposed in history and exact duplicate sources. Before reading a document, the Builder records the search event that revealed it. Document-identity checks and semantic validation together determine whether the required knowledge was already available in history.

\paragraph{Construction Models and Refinement}
All construction roles, including multi-condition enhancement, use GLM-5.3 Flash through Zhipu Coding Plan at temperature 0. The Profile builder uses a per-call output limit of 16,384 tokens. Question writing allows 8,192 tokens; the history-removal solver and final validator allow 24,576, and the requirement comparator allows 16,384. Separate prompts define each role's task and evidence access, as shown in Figure~\ref{fig:construction}.

\subsection{Acceptance and Evidence Mapping}
The history-removal test gives a blind solver the current question and new supporting text. A comparator checks its answer against the reference answer and necessary requirements. A task is labeled answerable without history when all requirements are satisfied, and history-dependent when a requirement is unmet because the historical context is missing. Remaining cases are labeled as having insufficient evidence. Final validation checks whether the task naturally extends the interaction, whether personal conditions are grounded in their sources, whether historical knowledge is insufficient, whether new evidence supports the answer, and whether the answer adds information and satisfies safety requirements. It retains the history-dependence label from the preceding test.

We represent each answer requirement as $e_j=(r_j,U_j,F_j,K_j,a_j)$, linking the requirement $r_j$ to verbatim user statements $U_j$, personal conditions $F_j$, knowledge rules $K_j$, and the required judgment or action $a_j$. Personal evidence traces back to user messages, and knowledge evidence traces back to documents. Unknown premises are handled through conditional branches. Programmatic checks validate identifiers and reference integrity, while a language model assesses semantic support against the source text.

\paragraph{Item Refinement and Enhancement}
\label{app:enhancement}
Refinement jointly updates the question, reference answer, requirements, and evidence mapping while preserving public history and personal conditions. Rewording retains the supporting document. Multi-condition enhancement combines at least two disclosed conditions around the initial task, identifying the judgment affected by each, and may replace the document with a retrieved candidate from the same Topic. All drafts undergo the history-removal test and final validation of source support and answerability; enhancement additionally requires dependence on multiple conditions. Failed drafts are excluded, and revisions are re-evaluated under the same criteria.
\section{Implementation and Experimental Settings}
\label{app:reproducibility}

\subsection{Memory and Retrieval Implementations}
\label{app:implementation}
Mem0 uses the non-graph memory path in the official SDK 2.0.20, with default extraction and update operations and local Qdrant storage. Mem0+Hybrid and MemKB share the resulting memories. For MemKB, an unresolved historical Read reference contributes no historical document link; memory records, valid historical links, and offline-neighbor links are retained.

Graphiti uses SDK 0.31 with Neo4j, stores each complete interaction pair as an episode, and retrieves relational facts with its built-in RRF. A-MEM uses the authors' implementation of note analysis, neighbor linking, memory evolution, and retrieval expansion. Each A-MEM memory response is capped at 5,120 tokens, including the serialized return structure. Notes are deduplicated and packed whole in their original retrieval order. All memory stores are built from the same 599 public interaction pairs, retaining message roles and sources.

The memory baselines expose separate \texttt{memory\_search} and \texttt{kb\_search} tools; MemKB uses \texttt{memkb\_search}. All methods share Read and the knowledge index, with ColaSearch's term-based keyword retrieval.

\subsection{Model and Retrieval Configuration}
\label{app:run-configuration}
Table~\ref{tab:run-configuration} lists model and retrieval settings, including the Qwen3 encoder~\cite{zhang2025qwen3}. Offline discovery retrieves 20 semantic candidates per card and excludes previously exposed documents. Each card lists retrieved neighbors followed by historical Read documents, preserving their respective orders; round-robin expansion deduplicates across cards.

All methods share the answering instructions for preserving actors, time, and unknown conditions and addressing each requested decision.

\begin{table*}[t]
\centering
\caption{Configuration for the main comparison. Output limits apply per call, and the agent budget applies per task.}
\label{tab:run-configuration}
\small
\begin{tabularx}{\textwidth}{@{}p{.19\textwidth}X@{}}
\toprule
Component & Configuration\\
\midrule
Answer model & DeepSeek v4.1 Flash; temperature 0; output limit 16,384 tokens; Agent budget 128,000 tokens. Mem0 and MemKB use OpenCode; No Memory, Graphiti, and A-MEM use SiliconFlow with OpenCode fallback.\\
Online interaction & Up to four tool rounds and one final-answer round; early completion allowed. A round is one model response and may contain multiple tool calls. Read: at most 100 lines and 3,000 tokens per call.\\
Accuracy evaluator & GLM-5.3 Flash; low reasoning effort; temperature 0; output limit 32,768 tokens. SiliconFlow and OpenCode, with identical model inputs and parameters.\\
Memory construction & Mem0 SDK extraction and update with GLM-5.3 Flash; Mem0 and MemKB share the resulting memories.\\
Embeddings & Qwen3-Embedding-0.6B for knowledge retrieval, Mem0, MemKB and Graphiti; all-MiniLM-L6-v2 for A-MEM memory.\\
MemKB state bank & GLM-5.3 Flash; output limit 16,384 tokens (32,768 for one Profile). Up to 30 states with subjects, time, unknown conditions and source-message IDs; provided before the first query and in search results.\\
Memory retrieval & Mem0 returns 20 results by default, with a configurable count. Graphiti and A-MEM use 5 results or root notes by default, up to 20; A-MEM expands linked notes within its 5,120-token return budget.\\
MemKB candidates & Top 20 memory cards; round-robin expansion to at most 20 unique linked documents. Semantic and keyword channels each return at most 20 candidates.\\
Fusion and output & Equal-weight RRF with $\kappa=60$; 5 documents by default, at most 20. Up to 20 deduplicated memories with source-card and message IDs.\\
\bottomrule
\end{tabularx}
\end{table*}

\subsection{Component Ablations}
\label{app:component-ablations}
The component ablations cover all 111 retained tasks (40, 36, and 35) from three Profiles. The full MemKB configuration reuses answers and judgments from the main comparison. The ablated configurations use the same models, evaluation criteria, and interaction budgets.

The state-summary ablation removes the state lists supplied before the first query and in search results, retaining query-relevant memory. The document-link ablation clears each card's linked document set while retaining memory vectors, the state summary, returned memory text, and the source-attribution constraint. Table~\ref{tab:native-link-ablation} reports the results.

\section{Supplementary Analysis and Cases}
\label{app:analysis-supplement}

\subsection{Paired Estimation}
\label{app:statistical-protocol}
The main comparison preserves the pairing between methods on all 349 tasks. We use a cluster bootstrap over the ten Profiles with 20,000 replicates and random seed 20261014. MemKB's Accuracy gain over Mem0+Hybrid is 6.30 percentage points, with a 95\% confidence interval of [3.19, 9.09].

\subsection{Task-Composition Sensitivity}
\label{app:composition-sensitivity}
Figure~\ref{fig:experiment-analysis}(d) stratifies tasks by Profile and the number of answer requirements, then adds construction cohort in a further adjustment. Each analysis retains strata containing both single-Fact and multi-Fact tasks. We calculate MemKB's Accuracy gain over Mem0+Hybrid for each group within each stratum, then aggregate the gains using stratum task counts as weights. Both groups receive the same stratum weights. The two common-support sets contain 316 and 275 tasks, respectively.
Construction cohort records the initial Item admission path: direct acceptance (155 tasks) or acceptance after revision and revalidation (194 tasks), as identified by the construction metadata.

Resampling preserves task-level pairings and recomputes stratum weights in each replicate, using the bootstrap settings above. The unadjusted gain difference between multi-Fact and single-Fact tasks is 4.39 percentage points, with a 95\% confidence interval of [$-5.52$, 15.20]. Adjusting for Profile and requirement count yields a difference of 9.54 points, with an interval of [$-2.21$, 21.75]; adding construction cohort yields 6.39 points, with an interval of [$-4.29$, 18.67]. The figure's error bars show confidence intervals for each group's paired gain. Intervals for the between-group difference are calculated by subtracting the gains within each bootstrap replicate. These exploratory intervals are unadjusted for multiple comparisons.

\subsection{Shared Background across Tasks}
\label{app:shared-background}
Grouping tasks by Profile and the exact set of selected Facts yields 98 groups across the 349 tasks. Of these, 60 groups involve at least two distinct reference documents and cover 311 tasks (89.1\%). Table~\ref{tab:state-rule-transfer} presents three tasks that share the same personal background. Each question calls for different rules, whose application depends on the relative's condition and the caregiver's distance recorded in the history.

\begin{table*}[t]
\centering
\caption{Three tasks sharing the same personal background: the user cares for a relative with Alzheimer's disease from a distance. Questions and answer points are paraphrased.}
\label{tab:state-rule-transfer}
\small
\begin{tabularx}{\textwidth}{@{}p{.24\textwidth}p{.27\textwidth}X@{}}
\toprule
Current question & Knowledge to retrieve & History-informed judgment\\
\midrule
Would my relative qualify for and be suited to a dementia-specific adult day program? & Target population and enrollment requirements. & Alzheimer's disease matches the target population; formal diagnostic documentation required for enrollment remains to be confirmed.\\
Does respite care apply, at home or during a short inpatient stay? & Hospice eligibility and respite-care arrangements. & The user is a family caregiver. If hospice enrollment is established, explain home and short-term inpatient options.\\
How should my relative prepare for a winter storm or power outage? & Dementia-specific supplies, communication and local assistance. & Adapt preparations to cognitive impairment and long-distance caregiving; identify local helpers and agree on responsibilities.\\
\bottomrule
\end{tabularx}
\end{table*}

\subsection{Case Evidence}
\label{app:cases}
\label{app:case-source}
\label{app:paired-diagnosis}
The HFC eligibility example in Table~\ref{tab:case} draws on the user's public statements about their caregiving relationship and living alone, together with the frozen \href{https://wearehfc.org/care-grants}{Care Grants} page. Caregiving role and co-residence each determine the applicability of the relevant rules.

\paragraph{Treatment History}
For task \path{item_0a49978ebac6}, user message \path{tu_0f0b8cae0621_user} states,  ``I'm a survivor of childhood cancer, and I still go for long-term follow-up care as an adult to keep an eye on how my treatment might be affecting me.'' Both methods read the NICHD document \path{kb_4cd3cbc8ac}; its opening 40 lines include the chemotherapy guidance and the rule for women ``undergoing radiation therapy''. Mem0+Hybrid answers, ``no record of HIV, HTLV, tuberculosis, cancer treatment, or radiation''. MemKB instead keeps treatment details conditional: ``I don't have your actual treatment details''.

\paragraph{Medication Class}
For task \path{item_21c9bf907c11}, message \path{tu_5415ad4050d9_user} states, ``I take prescription pain medicine that my doctor prescribed.'' Both methods read the CVS document \path{kb_web_73f82c20a758ffd8} in full (Read offsets 0--18), including the condition ``at increased risk for opioid overdose'' and access ``without an individual prescription''. MemKB writes, ``You do take a prescribed opioid pain medicine''; Mem0+Hybrid writes, ``You're currently taking a prescription opioid pain medicine''.
\paragraph{Error-Analysis Evidence}
In task \path{item_a6b3ffbe02ec}, user message \path{tu_5add4071b6e7_user} states, ``I take prescribed pain medicine for my cancer-related pain''. State \path{state_006} adds ``(opioid)'', and the answer consequently treats opioid-related naloxone guidance as applicable. The FDA document \path{kb_web_cf4e473771cdc3d7} supplies the opioid-use condition; the historical statement leaves the medication class unspecified.

In the scalp-cooling task \path{item_0b65cfc2b717}, document \path{kb_web_838a73aeafde37e1} states that cooling is not recommended for ``migraine headaches from cold temperatures''. The returned snippet mentions cancer-related exclusions but omits this rule; the agent instead reads \path{kb_web_a5e06a9cd52e5f04}. In the neuropathy task \path{item_b58978a86ea7}, the agent reads \path{kb_web_f0f60132da209c71}, whose heat-risk list includes ``warm foot baths'', yet recommends a warm foot bath with temperature checks.

\section{Evaluation Protocol}
\label{app:evaluation-prompts}

\subsection{Metrics and Scoring Inputs}
\label{app:metric-details}
\label{app:strict-accuracy-prompts}
Accuracy follows Section~\ref{sec:evaluation}: one call checks personal facts, knowledge claims, and task requirements, returning \texttt{pass}, \texttt{fail}, or \texttt{needs\_adjudication}. All main-comparison and ablation judgments resolve to \texttt{pass} or \texttt{fail}; Accuracy divides passes by all tasks in the comparison. The 48 MemKB errors in Section~\ref{sec:error-analysis} are failed answers.

Read Recall follows Section~\ref{sec:evaluation}, using unique reference documents with nonempty Read returns; search snippets, failed reads, and empty reads do not count. First-search and overall search hit rates measure reference-document presence in the first or any successful knowledge-search response, respectively. Their denominator includes all cohort tasks, counting tasks without successful searches as misses. Memory-only searches, failed calls, and internal candidates are excluded.

All methods share the task's complete public history, reference answer, requirements, and fixed historical and new-support documents as evaluation references. The 1,015 requirements specify applicability conditions and unknown premises. Candidate inputs add the answer and actual Search snippets and Read text. Agent-visible evidence includes supplied states, memories, and Search/Read returns; internal candidates and evaluator-only material are excluded. Full construction, memory, and evaluation prompts and input templates will be provided on GitHub. Model parameters appear in Appendix~\ref{app:run-configuration}.

\subsection{Answer-Quality Dimensions}
\label{app:answer-dimensions}
We evaluate the 222 Mem0+Hybrid and MemKB answers on the 111-task ablation cohort. State Fidelity checks whether represented personal conditions preserve the historical speaker, actor, timing, and uncertainty. Condition Coverage measures whether the personal premises needed for the current task are substantively reflected. Evidence Grounding assesses support for knowledge claims in the supplied documents and actual retrieved passages. Rule Application evaluates whether prerequisites, restrictions, exceptions, and timing are correctly applied to the user's situation. Answer Completeness measures coverage of requested subquestions and necessary actions. Each call returns a score with candidate excerpts, source evidence, and a rationale.

\subsection{Error Categories and Evidence Visibility}
\label{app:error-diagnosis}
The primary category identifies each failed answer's decisive error; its Accuracy verdict is unchanged. \emph{Personal Assumptions} covers treating unknown states, plans, advice, or another person's information as established personal facts. \emph{Context Underuse} covers failing to apply established user conditions to the current judgment. \emph{Rule Misapplication} covers incorrect knowledge claims, rule interpretations, or applicability judgments. \emph{Answer Omissions} covers unanswered necessary subquestions, actions, restrictions, or steps when no more direct personal-context error explains the omission. \emph{Other Issues} covers errors outside these categories or unresolved attribution and source conflicts. Up to two secondary categories record other independently supported errors; Figure~\ref{fig:error-analysis} reports primary categories.

Visibility concerns the personal premises and rules needed to resolve the errors, assessed against the agent's question, states, memories, and Search and Read returns. \emph{Visible} indicates that all required evidence was available; \emph{Partly visible}, that some evidence was available but key content was missing; and \emph{Not visible}, that the key evidence was absent. Evaluator-only material and unreturned candidates are excluded; for unsupported personal assertions, visibility concerns the correct condition or uncertainty. \emph{Uncertain} is used when the records or source conflicts prevent a reliable judgment; no answer received this label.

\end{document}